\documentclass[letterpaper, 10pt, conference]{ieeeconf}
\IEEEoverridecommandlockouts
\usepackage{cite}
\usepackage{amsmath,amssymb,amsfonts}
\usepackage{algorithmic}
\usepackage{graphicx}
\usepackage{textcomp}
\usepackage{xcolor}
\usepackage[margin=20 mm]{geometry}
\usepackage{tikz}

\def\BibTeX{{\rm B\kern-.05em{\sc i\kern-.025em b}\kern-.08em
    T\kern-.1667em\lower.7ex\hbox{E}\kern-.125emX}}

\usepackage{parskip}
\usepackage{multicol}
\usepackage[pdfa]{hyperref}
\usepackage{multirow}
\usepackage{booktabs}
\usepackage{tabularx}
\usepackage{bm}
\usepackage{bbold}
\usepackage{comment}
\usepackage[font=small]{caption}
\usepackage[font=footnotesize]{subcaption}

\usepackage[acronym]{glossaries}
\newacronym{ransac}{RANSAC}{Random Sample Consensus}

\usepackage{balance}

\begin{document}

\vspace{100cm}
\title{
\LARGE 
\bf Plane detection and ranking via model information optimisation\\
\thanks{This research is supported by the Home Team Science and Technology Agency (HTX).}
}
\author{Daoxin Zhong$^{1}$, Jun Li$^{1}$, Meng Yee (Michael) Chuah$^{1}$ 
\thanks{$^{1}$Authors are with the Institute for Infocomm Research (I\textsuperscript{2}R), A*STAR, Singapore.
\href{michael_chuah@i2r.a-star.edu.sg}{\tt\small{zhongdx, jli, michael$\_$chuah@i2r.a-star.edu.sg}}
}
}
\maketitle



\begin{abstract}
Plane detection from depth images is a crucial subtask with broad robotic applications, often accomplished by iterative methods such as Random Sample Consensus (RANSAC). While RANSAC is a robust strategy with strong probabilistic guarantees, the ambiguity of its inlier threshold criterion makes it susceptible to false positive plane detections. This issue is particularly prevalent in complex real-world scenes, where the true number of planes is unknown and multiple planes coexist. In this paper, we aim to address this limitation by proposing a generalised framework for plane detection based on model information optimization. Building on previous works, we treat the observed depth readings as discrete random variables, with their probability distributions constrained by the ground truth planes. Various models containing different candidate plane constraints are then generated through repeated random sub-sampling to explain our observations. By incorporating the physics and noise model of the depth sensor, we can calculate the information for each model, and the model with the least information is accepted as the most likely ground truth. This information optimization process serves as an objective mechanism for determining the true number of planes and preventing false positive detections. Additionally, the quality of each detected plane can be ranked by summing the information reduction of inlier points for each plane. We validate these properties through experiments with synthetic data and find that our algorithm estimates plane parameters more accurately compared to the default Open3D RANSAC plane segmentation. Furthermore, we accelerate our algorithm by partitioning the depth map using neural network segmentation, which enhances its ability to generate more realistic plane parameters in real-world data.
\end{abstract}

\section{Introduction}
\begin{figure}[t]
\captionsetup[subfigure]{labelformat = empty}
    \centering
    \begin{subfigure}{0.23\textwidth}
        \centering
        \includegraphics[width=\textwidth]{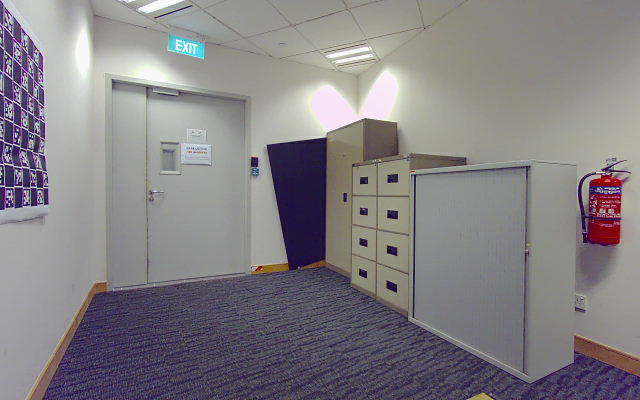}
        \caption{RGB Image}
    \end{subfigure}
    \begin{subfigure}{0.23\textwidth}
        \centering
        \includegraphics[width=\textwidth]{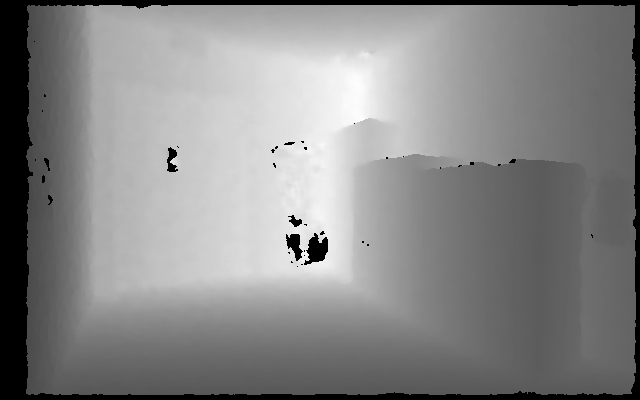}
        \caption{Depth Image}
    \end{subfigure}
    \begin{subfigure}{0.23\textwidth}
        \centering
        \includegraphics[width=\textwidth]{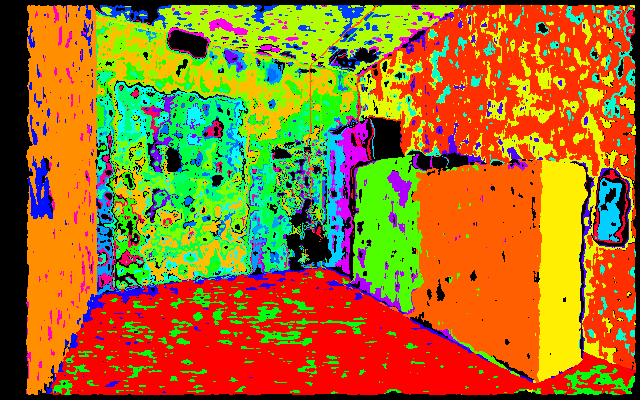}
        \caption{Open3D Plane Detection}
    \end{subfigure}
    \begin{subfigure}{0.23\textwidth}
        \centering
        \includegraphics[width=\textwidth]{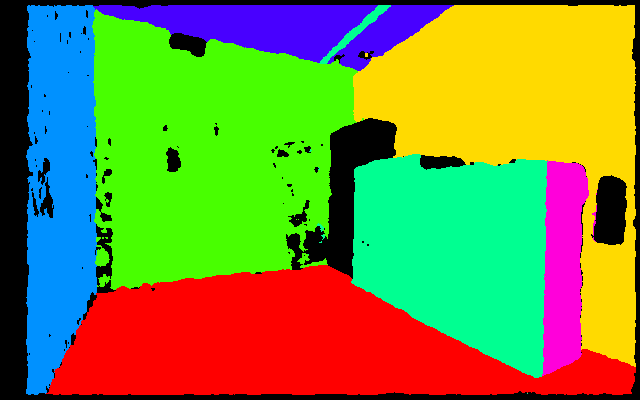}
        \caption{Our Approach}
    \end{subfigure}
    \caption{The traditional RANSAC approach to plane detection requires human turning to identify the optimal inlier ratio. Otherwise, the algorithm may generate many additional false positive planes, resulting in a fragmented plane mask as seen here. This is unnecessary in our approach as the algorithm, it leverages model information to determine ground truth model. Thus, if supplied with an accurate noise model of the scene, the algorithm pauses when the true planes are found and rejects all remaining points as noise.}
    \label{fig:intro}
\end{figure}
The prevalence of flat surfaces in human-centric environments makes plane estimation a critical step for many robotics-related tasks. Legged robots rely on floor plane estimation to guide foot placement accurately during navigation \cite{Woo2020}. When surveying buildings using mobile mapping systems, identifying plane intersections is an important intermediate step in generating a twin CAD model \cite{Schmidt2023}. In the domain of monocular depth estimation, plane constraints can serve as valuable priors, frequently incorporated into the design of loss functions during model training \cite{shao2023nddepth}\cite{PlaneDepth}\cite{shi2023planerectr}.

Physical plane parameters are typically estimated from point cloud data, captured via depth sensors such as RGB-D cameras or LiDARs. For many years, \gls{ransac} has been a popular choice for this task \cite{Woo2020}\cite{Schmidt2023}, owing to its robustness and strong probabilistic guarantees. However, RANSAC performs best when finding individual plane features and struggles when applied to real-world scenes with multiple planes. While it is possible to extract each individual plane consecutively via RANSAC, there is no clear stopping point to this process without prior knowledge of the true plane count, thus allowing for the detection of false positive planes. RANSAC also selects inliers based on an arbitrary threshold that requires manual tuning. Valid planes may become fragmented due to sensor noise if the criterion is too strict, while distinct planes may be grouped together if the criterion is too lenient (See Fig \ref{fig:intro}). Recent deep learning techniques have incorporated RGB image data to improve the quality of the detected planes. However, the ambiguities mentioned above also affect the ground-truth data used to train these models, hence limiting their efficacy.

Inspired by the work of \textit{Yang and F{\"o}rstner} \cite{yang2010plane}\cite{Forestner_original}, we have decided to approach these problems through the lens of information theory. Since RGB-D cameras and LiDARs are digital devices, their depth readings take on discrete values within the sensor's operating range. Treating each observation as a random variable, the existence of a plane constraint limits the distribution of its inliers, hence lowering the overall model information. Thus, by generating a list of candidate planes via repeated random sub-sampling, the optimal best fit plane can be found via model information minimisation. An assignment mask may be used to penalise models with too many planes, allowing the process to be repeated consecutively to determine the true plane count and their parameters. Our work aims to correct the mistakes found in \cite{yang2010plane} and incorporate additional constraints using the physics of the depth sensor. By calculating the information reduction of each plane, it also provides a clear metric to rank the quality of each detected plane. We experimented with the utility of neural network-based image segmentation techniques that can improve the run time of this algorithm by first partitioning the depth map based on semantics. Ultimately, we hope to provide a generalised framework to generate accurate plane parameters that are useful for other downstream robotic applications.

In summary, our paper aims to address the problem of plane detection through the lens of model information optimisation. Given a set of digital depth data, captured by some RGB-D camera or LiDAR, our proposed algorithm 
\begin{enumerate}
    \item finds the exact number of planes and their parameters that exists in the point cloud given some sensor noise model.
    \item finds the ranking of each recovered plane based on their respective information reduction.
    \item can be accelerated by partitioning the depth map based on its semantic mask
\end{enumerate}
An implementation of our algorithm can be found at \href{https://github.com/mcx-lab/InformationOptimisation}{\textcolor{blue}{https://github.com/mcx-lab/InformationOptimisation}}.

\section{Background}

\textbf{RANSAC:} 
First proposed by \cite{Fischler1981RandomSC} in 1981, RANSAC remains a popular strategy for algorithmic plane detection \cite{Woo2020}\cite{Schmidt2023}. By evaluating the number of inliers for a list of candidate solutions through repeated random sub-sampling, the optimal solution may be found within a certain confidence level. This fact is backed by strong probabilistic guarantees which state that the required number of trial solutions depends only on the confidence level, the inlier ratio and the model parameters. This fact, together with its ease of implementation, is why both \cite{yang2010plane} and our work have chosen to leverage repeated random sub-sampling for candidate plane generation. However, the utility of RANSAC for plane detection in real-world applications is limited. First, the algorithm tends to converge slowly, leading to long algorithmic run-times in complex scenes, where the inlier ratio for each plane may be small. Additionally, RANSAC struggles with poor noise tolerance. While this issue can be mitigated through additional data processing techniques, such as error thresholding \cite{ransac_thresholding}, the noisy data affects the accuracy of the detected plane parameters and may lead to false positive plane detection.

To mitigate these problems, a model verification step may be adopted to eliminate degenerate models. \cite{Raguram2013} attempts to provide a generalised framework for RANSAC that evaluates the proposed plane parameters based on qualitative metrics, such as the Bail-out and SPRT test. Similarly, \cite{yang2010plane} leverages the discrete nature of digital depth data to measure the informational likelihood of each model. In contrast to the arbitrary threshold used in RANSAC for inlier assignment, their method assigns points based on their contribution to the overall model information, thus eliminating the bias and subjectivity involved in the tuning process. We aim to build on this idea by further integrating the sensor’s physics and noise level to better estimate the information of each proposed model.

\textbf{Other methods:}
To overcome the slow convergence of repeated random sub-sampling solutions, some papers have instead opted for localised methods of plane estimation. This typically involves a normal estimation step that aims to estimate the normal vector and scalar plane distance of each point. They are then grouped into planes based on their normal similarity, through flood-filling \cite{Hwang2023}, Felzenszwalb segmentation \cite{shao2023nddepth} or other graph-based methods \cite{graph_1}\cite{graph_2}. We experimented with some of these ideas in our work to improve the accuracy of our detected plane along plane intersection, but found that they only led to unsatisfactory results. (See Section \ref{Section: Limitations}).

Recently, more attention has been dedicated to neural network based plane estimation methods \cite{Jin2018}\cite{PlaneRecTR++}. These models usually accept an RGB image as input, using it alone to estimate the plane parameters without the need for any depth data. Note that while our experiments included the use of RGB images for semantic segmentation, this was solely aimed at speeding up our algorithm run time by creating smaller partitions in our depth images. The bulk of our work remains focused on the theoretical formulation of plane model information, which estimates plane parameters based solely on depth data. These methods are instead included in our paper to benchmark the accuracy of our estimated planes.

\section{Methods}
\subsection{Mathematical Formulation}
\label{methods:mathematic_formulation}

To formalise the plane detection problem, let $X$ be a set of $k$ data points in $\mathbb{R}^n$, where each point $\bm{x_i}$ lies along discrete intervals $\epsilon$ between $[0,R]$, $X = \{\bm{x_1}, \bm{x_2}, \dots, \bm{x_k}\ | \bm{x_i^j} \in \{0,\epsilon,2\epsilon,\dots,R\}, 1\leq i\leq k,0\leq j \leq n\}$. Our goal is to determine the number of hyperplane constraints that exist in $X$ and assign the points to the planes. In this framework, each proposed number of hyperplanes and their respective hyperplane parameters represent a candidate model. Our goal is to evaluate the information of the model and select the one with the least information as the ground truth, since minimizing the model's information complexity also maximizes its likelihood \cite{Forestner_original}.

We begin by considering the information of just two simple models: $O$ (no plane constraint found, all outliers) and $1P+O$ (one plane constraint found, rest outliers). In the $O$ case, each point is considered noise and modeled to be independent and identically distributed (i.i.d.). The information of model $O$ is therefore
\begin{equation}
\begin{split}
\Phi_0 & = I(X|O) = - \ln P(X|O) \\
& = -\ln \prod^k_{i=1} P(\bm{x_i}|R,\epsilon) = -\sum^k_{i=1}\ln P(\bm{x_i}|R, \epsilon) \\ 
& = -\sum^k_{i=1} \ln (1/(R/\epsilon)^n)
= kn\ln(R/\epsilon)
\end{split}
\end{equation}
To evaluate the information of the $1P+O$ model, assuming that there exist $k_0$ points of noise and $k_1$ points on the hyperplane. The first $k_0$ points require $k_0n\ln (R/\epsilon)$ nats of information, similar to the $O$ model. A hyperplane will constrain the remaining $k_1$ points. It is parametrised by a normal vector $\bm{n}$ and scalar distance $d$ that are uniquely determined by $n$ distinct points, hence requiring $n\ln(R/\epsilon)$ nats of information. Each $\bm{x_i}$ constrained by the hyperplane can be split into its two orthogonal components, one parallel $\bm{x_{i,p}}$ and one normal $\bm{x_{i,n}}$ to the plane. $\bm{x_{i,p}}$ can be mapped onto any valid interval on the hyperplane, hence requiring $k_1(n-1)\ln(R/\epsilon)$ nats of information. $\bm{x_{i,n}}$ will be non-zero due to some error, $\delta$, between the point and the hyperplane where
\begin{equation}
\delta(\bm{x_i}) = \bm{n} \cdot \bm{x_i} + d = \bm{n} \cdot \bm{x_{i,n}} + d
\end{equation}
This error is assumed to fall within a zero-mean normal distribution with variance $\sigma^2$, $\delta(\bm{x_i}) \sim N(0, \sigma^2)$. 

In summary, to quantify the information of each $k_1$ point
\begin{equation}
\begin{split}
& P(\bm{x_i}|1P+O) = P(\bm{x_i}|\bm{n},d,R,\epsilon) \\
& = P(\bm{x_{i,p}},\bm{x_{i,n}}|\bm{n},d,R,\epsilon) = P(\bm{x_{i,p}}|R,\epsilon)  P(\bm{x_{i,n}}|\bm{n},d,\epsilon) \\
& = \frac{1}{(R/\epsilon)^{n-1}} * \\
& \qquad \int^{x_{i,n}+\frac{\epsilon}{2}}_{x_{i,n}-\frac{\epsilon}{2}}\frac{1}{\sqrt{2\pi\sigma^2}} \exp (-\frac{1}{2} \frac{(\bm{n}\cdot\bm{\bar{x}_{i,n}}+d)^2}{\sigma^2}) d\bar{x}_{i,n}
\\
& \approx \frac{1}{(R/\epsilon)^{n-1}} \frac{\epsilon}{\sqrt{2\pi\sigma^2}} \exp (-\frac{1}{2} \frac{(\bm{n}\cdot\bm{x_{i}}+d)^2}{\sigma^2})  \qquad \bigl[\epsilon\ll \sigma\bigr]
\end{split}
\end{equation}

\begin{equation}
\begin{split}
I(\bm{x_i}|1P+O) 
&= (n-1)\ln (R/\epsilon) \\
& + \frac{1}{2} \frac{(\bm{n}\cdot\bm{x_i}+d)^2}{\sigma^2} + \frac{1}{2} \ln(2\pi\sigma^2/\epsilon^2)
\end{split}
\end{equation}

Thus, the total information of the model is
\begin{equation}
\begin{aligned}
\Phi_1 &= I(X|1P+O) \\
       &= k \ln 2 
       + k_0 n \ln (R/\epsilon) \\
       &\hspace{3.6em} + n \ln (R/\epsilon) + k_1 (n-1) \ln (R/\epsilon) \\
       &\hspace{3.6em} + \sum^{k_1}_{i=1} \left[\frac{1}{2} \frac{(\bm{n} \cdot \bm{x_i}+d)^2}{\sigma^2} + \frac{1}{2} \ln \left( 2\pi \sigma^2 / \epsilon^2 \right)\right]
\end{aligned}
\end{equation}

Note the inclusion of the leading $k\ln2$ term. This represents the information of the assignment mask, which assigns each point as an outlier or to the hyperplane.

Generalizing this to a model with $N$ hyperplanes, the total information of the model is
\begin{equation}
\begin{split}
\Phi_N 
&= I(X|NP+O) \\
&= k\ln (N+1) + k_0n\ln (R/\epsilon) \\
&\hspace{4em}+\sum^N_{i=1} \Bigl[
n\ln(R/\epsilon) 
+ k_i(n-1)\ln (R/\epsilon) \\
&\hspace{4em} + \sum^{k_i}_{j=1} \bigl[\frac{1}{2} \frac{(\bm{n_i}\cdot\bm{x_j}+d_i)^2}{\sigma^2} + \frac{1}{2} \ln(2\pi\sigma^2/\epsilon^2)
\bigr]\Bigr]
\end{split}
\end{equation}
such that $\sum_{i=0}^N k_i = k$.

Setting $n=3$ when searching for planes within 3D point cloud data, our model information formula corrects two mistakes found in \cite{yang2010plane}. First, the information of the assignment mask increases with the number of planes, penalizing model complexity to prevent overfitting. Furthermore, the information due to the normal error decreases by $\ln \epsilon$ instead of $\ln \epsilon^n$, since each plane constraint only limits the normal component of $\bm{x_i}$, which is one-dimensional. This also keeps the overall information reduction via discretisation invariant at $n\ln\epsilon$ per point.

Formalising the assignment mask for each point in $X$: Given a dataset $X$ with $k$ points, find the optimal number of planes $N^\star$ and the assignment mask, $M^\star=\{m_1^\star, m_2^\star, \dots,m_k^\star|m_i^\star \in \{0,...,N^\star\}\}$, where each point is assigned to a plane index ($m_i\in\{1,2,\dots,N^\star\})$ or as an outlier $(m_i=0)$. $N^\star$ and $M^\star$ are parameters that minimises the following optimisation problem
\begin{equation}
\begin{split}
&\Phi_{N^\star,M^\star} 
= \min_{N,M} \Phi_{N,M} \\
&= \min_{N,M,\bm{n},d}
k\ln (N+1) + \sum^k_i \mathbb{1}_0(m_i)3\ln (R/\epsilon) \\
&\hspace{4em}+ \sum^N_{i=1} \biggl[
3\ln(R/\epsilon) +
\sum^{k}_{j=1} \mathbb{1}_i(m_j) \Bigl[2\ln (R/\epsilon)  \\
&\hspace{4em} + \frac{1}{2}\frac{(\bm{n_i}\cdot\bm{x_j}+d_i)^2}{\sigma^2} + \frac{1}{2} \ln(2\pi\sigma^2/\epsilon^2)
\Bigr]
\biggr]
\end{split}
\end{equation}

Solving for $M^\star$ becomes computationally more expensive as $N$ increases, since it is difficult to estimate the parameters of all $N$ planes concurrently. Therefore, we adopt the strategy suggested in \cite{yang2010plane} and employ repeated random sub-sampling to estimate each plane consecutively. Let $M_N$ be the best assignment mask found for a model with $N$ planes such that $M_0 = \{0, 0,\dots, 0\}$. The assignment mask for the first $N-1$ planes is kept constant ($M_N^i =M_{N-1}^i$ if $M_{N-1}^i \neq 0$), while a number of candidate planes is generated based on the remaining unselected points. For each candidate plane, $\bm{x_i}$ will be considered part of the $N$th plane ($M_N^i =N$) if $-\ln (R/\epsilon) + \frac{1}{2} \delta^2(\bm{x_i})/\sigma^2 + \frac{1}{2} \ln(2\pi\sigma^2/\epsilon^2) < 0$ or an outlier otherwise ($M_N^i =0$). The candidate plane with the most negative total information reduction is assumed to be the most likely $N$th plane. This algorithm is repeated up to some large $N$ to find the model with the most negative information, $\Phi_{\bar{N},\bar{M}_N}$, which is accepted as the most likely ground truth. Hence, we can conclude that there are $\bar{N}$ planes in the point cloud with the corresponding assignment mask $\bar{M}$.
\begin{equation}
\begin{split}
\Phi_{0,\bar{M}_0} &= 3k\ln(R/\epsilon)\\
\Phi_{N,\bar{M}_N} &= \Phi_{N-1,\bar{M}_{N-1}} + k \ln \frac{N+1}{N} + 3\ln(R/\epsilon) \\
&+ \min_{M_N,\bm{n_N},d_N} \sum^k_{i=1} \mathbb{1}_N(m_j) \Bigl[-\ln (R/\epsilon) \\
&\hspace{4em}+ \frac{1}{2} \frac{(\bm{n_N}\cdot\bm{x_i}+d_N)^2}{\sigma^2} + \frac{1}{2} \ln(2\pi\sigma^2/\epsilon^2) \Bigr] \\
\bar{N} &= \arg \min_N \Phi_{N,\bar{M}_N}
\end{split}
\label{eqn:practical_optimisation}
\end{equation}

\subsection{Physical Model of Depth Image}
\begin{figure}[h]
    \centering
    \includegraphics[width=\linewidth]{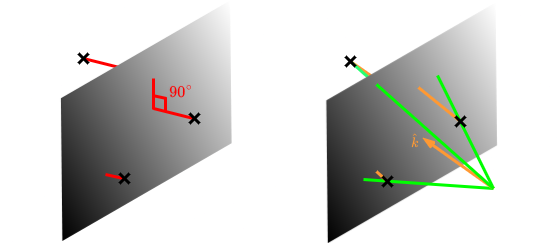}
    \caption{In the generalised plane detection problem, $\delta(\bm{x_i})$ is equals to the normal distance between $\bm{x_i}$ and the plane (red). However, when the point cloud is calculated through the depth image, $\bm{x_i}$ must lie on the image's pixel projection rays (green). Therefore, $\delta(\bm{x_i})$ is the parallel distance of $\bm{x_i}$ to the plane along the z-axis (orange).}
    \label{fig:drawing_of_error}
\end{figure}
Suppose that the point cloud is generated via a depth image captured by an RGB-D camera or LiDAR, $\bm{x_i}$ is extrapolated via the projection ray of each pixel. Since the point must lie on the projection ray, this serves as an additional linear constraint that reduces the information of each point from $3\ln(R/\epsilon)$ to $\ln(R/\epsilon)$. It is also no longer meaningful to calculate $\delta$ based on the normal distance of $\bm{x_i}$ from the plane, as the error due to sensor noise is applied strictly along the z-axis. Hence, $\delta$ is now the distance between $\bm{x_i}$ and the plane along the z-axis (See Fig \ref{fig:drawing_of_error}). Given the direction vector of the projection ray $\bm{d_i}$ and normal vector $\bm{n}$ and scalar distance $d$ of the plane, the new error is
\begin{equation}
\begin{split}
\bm{\bar{x}_i} &= \lambda \bm{d_i}, \lambda \in \mathbb{R} \\
\bm{n} \cdot \bm{\bar{x}_i} + d &= 0 \\
\bm{\bar{x}_i} &= -\frac{d}{\bm{n}\cdot\bm{d_i}}\bm{d_i} \\
\delta(\bm{x_i}) &= ( \bm{x_i} + \frac{d}{\bm{n}\cdot\bm{d_i}}\bm{d_i} ) \cdot \bm{\hat{k}}
\end{split}
\end{equation}

For many real-world sensors, the noise of the measured depth worsens at larger distances \cite{gemini2xl,d435,os1}. We can capture this fact in our model by introducing a depth-dependent sensor noise, $\sigma(\bm{x_i})$. This can be incorporated with Eqn \ref{eqn:practical_optimisation} to give the new information optimisation problem. 
\begin{equation}
\begin{split}
\Phi_{0,\bar{M}_0} &= k\ln(R/\epsilon)\\
\Phi_{N,\bar{M}_N} &= \Phi_{N-1,\bar{M}_{N-1}} + k \ln \frac{N+1}{N} + 3\ln(R/\epsilon) \\
&+ \min_{M_N,\bm{n_N},d_N} \sum^k_{i=1} \mathbb{1}_N(m_j) \Bigl[-\ln (R/\epsilon) \\
&\hspace{2.5em}+ \frac{1}{2} \frac{(( \bm{x_i} + \frac{d}{\bm{n}\cdot\bm{d_i}}\bm{d_i} ) \cdot \bm{\hat{k}})^2}{\sigma(\bm{x_i})^2} + \frac{1}{2} \ln\frac{2\pi\sigma(\bm{x_i})^2}{\epsilon^2} \Bigr] \\
\bar{N} &= \arg \min_N \Phi_{N,\bar{M}_N}
\end{split}
\end{equation}

\subsection{Semantic Partition}

If $r$ is the inlier ratio, at least $\ln(1-c)/\ln(1-r^3)$ candidates plane must be tested before a minimum confidence level of $c$ can be reached using repeated random sub-sampling\cite{Fischler1981RandomSC}. When $r$ is small and the total number of unassigned points $k$ is large, this can become computationally expensive. To address this issue, \cite{yang2010plane} suggests partitioning the point cloud into rectangular blocks and restricting the plane search to each partition. To modernise this approach, we instead choose to partition the point cloud using recent image segmentation techniques. Assuming the depth map is fused with an RGB image, the semantic masks produced by an image segmentation model can be leveraged to partition the corresponding point cloud. This strategy speeds up the overall algorithm run time, as it decreases both the values of $k$ $r$ for each partition, since the inlier ratio for the plane is expected to be higher within each semantic object.

Once all valid planes are found, the total information reduction of each plane can be calculated as $\sum^{k}_{j=1}[-\ln (R/\epsilon) + \frac{1}{2}\delta(\bm{x_i})^2/\sigma(\bm{x_i})^2 + \frac{1}{2} \ln(2\pi\sigma^2/\epsilon^2)]$. This is the metric that allows us to quantify the quality of each detected plane. Since a better-fitting plane will include points with a smaller error $\delta$, the best plane candidate is expected to have the most negative total information reduction. By ranking the planes based on this metric, poorer plane candidates may be eliminated or combined with other planes, giving a final, ordered list of detected planes.

\section{Experiments}
\subsection{Synthetic Data}

\begin{figure}[!tbhp]
\captionsetup[subfigure]{labelformat = empty}
    \centering
    \begin{subfigure}{0.22\textwidth}
        \centering
        \includegraphics[width=\textwidth]{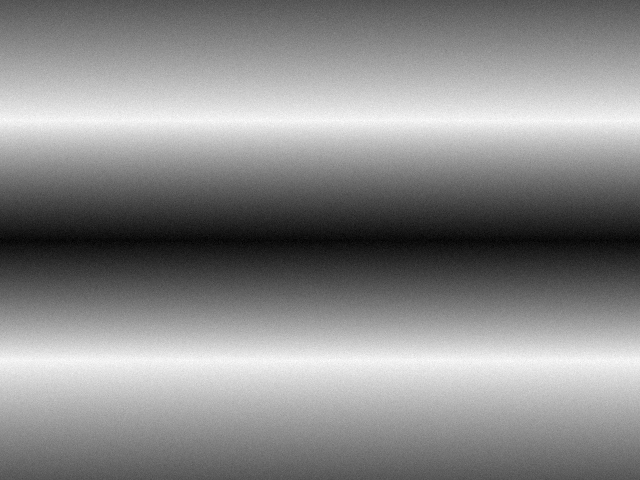}
    \end{subfigure}
    \begin{subfigure}{0.22\textwidth}
        \centering
        \includegraphics[width=\textwidth]{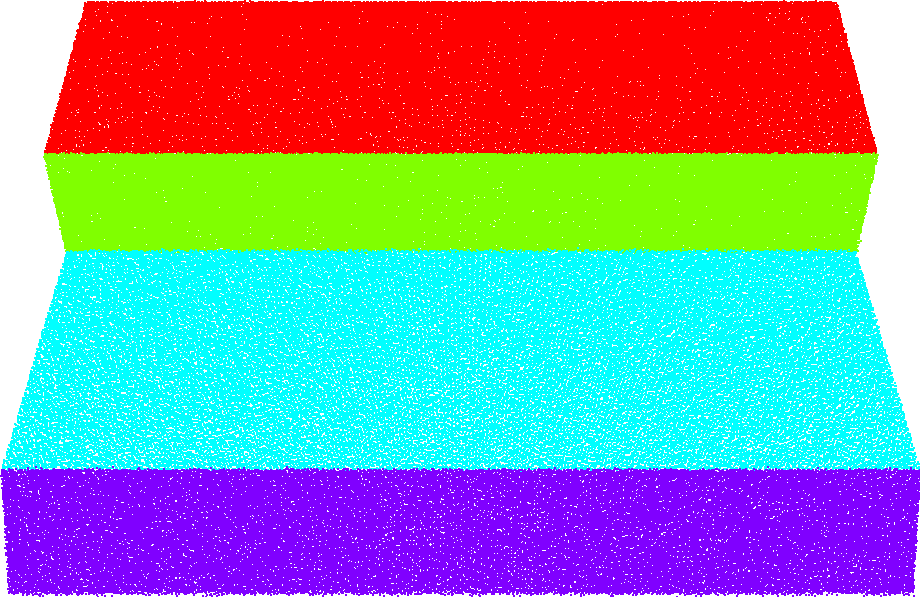}
    \end{subfigure}
    \begin{subfigure}{0.22\textwidth}
        \centering
        \includegraphics[width=\textwidth]{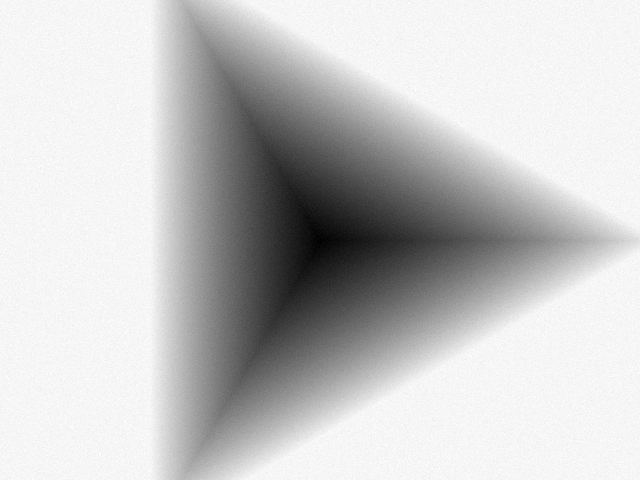}
        \caption{Depth Image}
    \end{subfigure}
    \begin{subfigure}{0.22\textwidth}
        \centering
        \includegraphics[width=\textwidth]{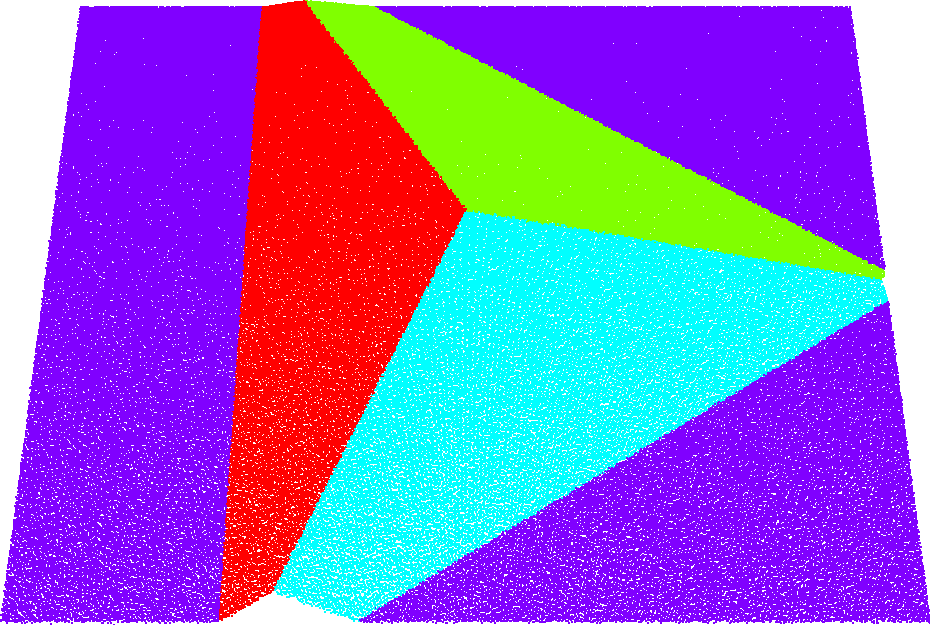}
        \caption{Plane Mask}
    \end{subfigure}
    \caption{Depth image and plane mask of our two synthetic test cases. Colours are used to differentiate the different planes.}
    \label{fig:Synthatic_corner_staircase}
\end{figure}

\begin{figure}[!tbhp]
\captionsetup[subfigure]{labelformat = empty}
    \centering
    \begin{subfigure}{0.22\textwidth}
        \centering
        \includegraphics[width=\textwidth]{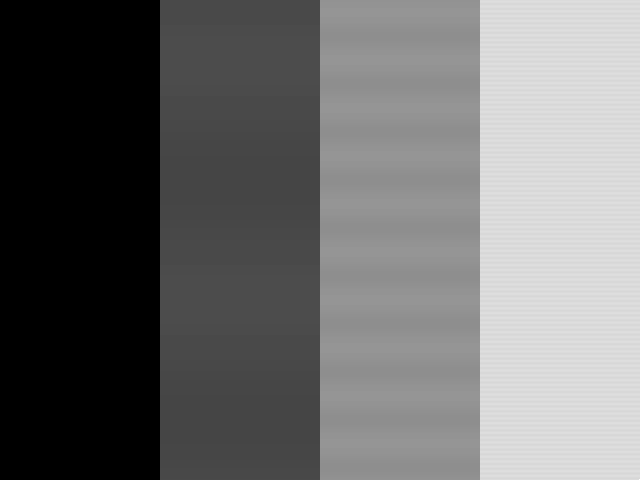}
        \caption{Depth Image}
    \end{subfigure}
    \begin{subfigure}{0.22\textwidth}
        \centering
        \includegraphics[width=\textwidth]{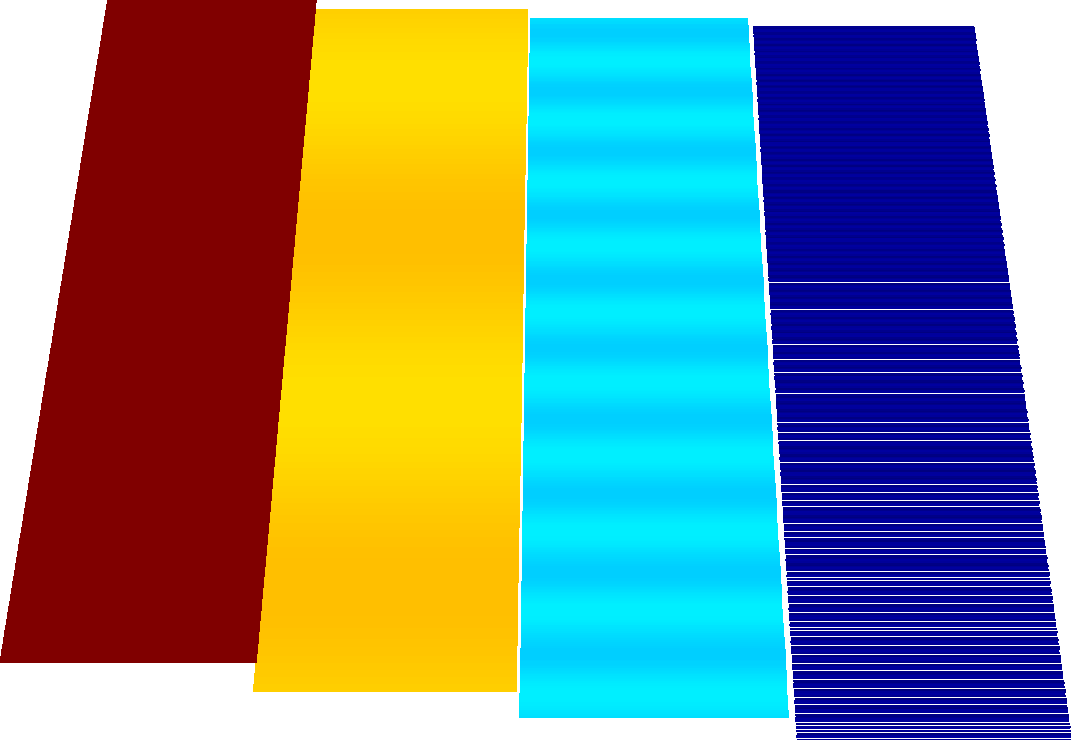}
        \caption{GT Mask}
    \end{subfigure}
    \begin{subfigure}{0.22\textwidth}
        \centering
        \includegraphics[width=\textwidth]{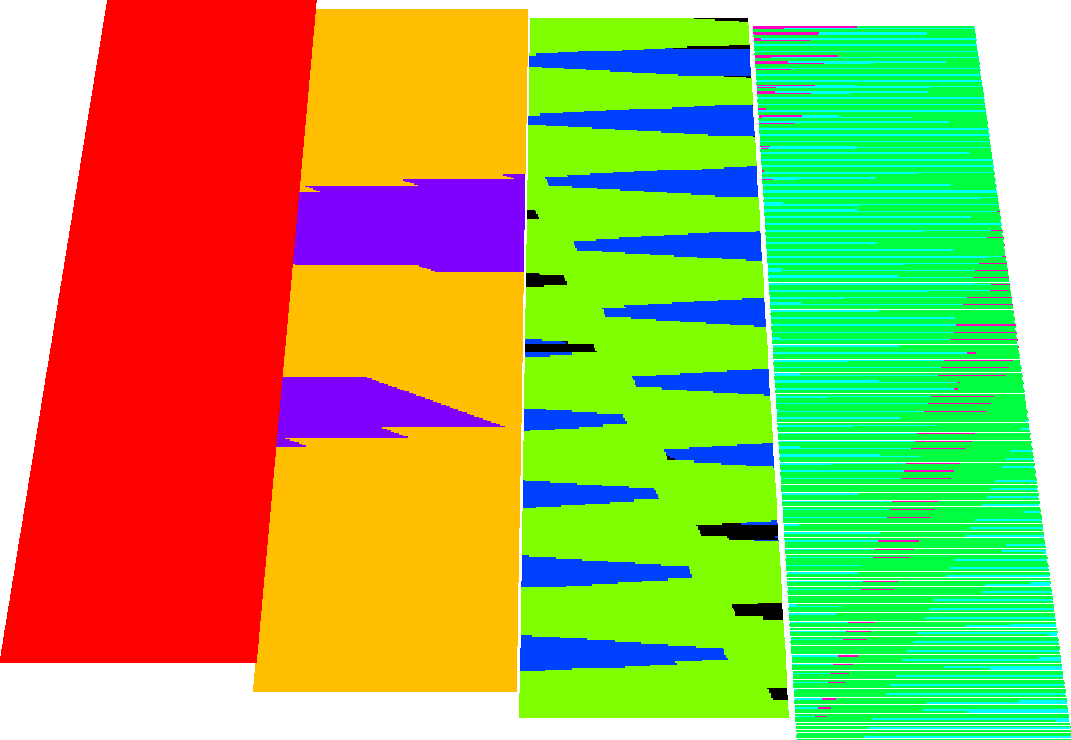}
        \caption{Open3d Planes}
    \end{subfigure}
    \begin{subfigure}{0.22\textwidth}
        \centering
        \includegraphics[width=\textwidth]{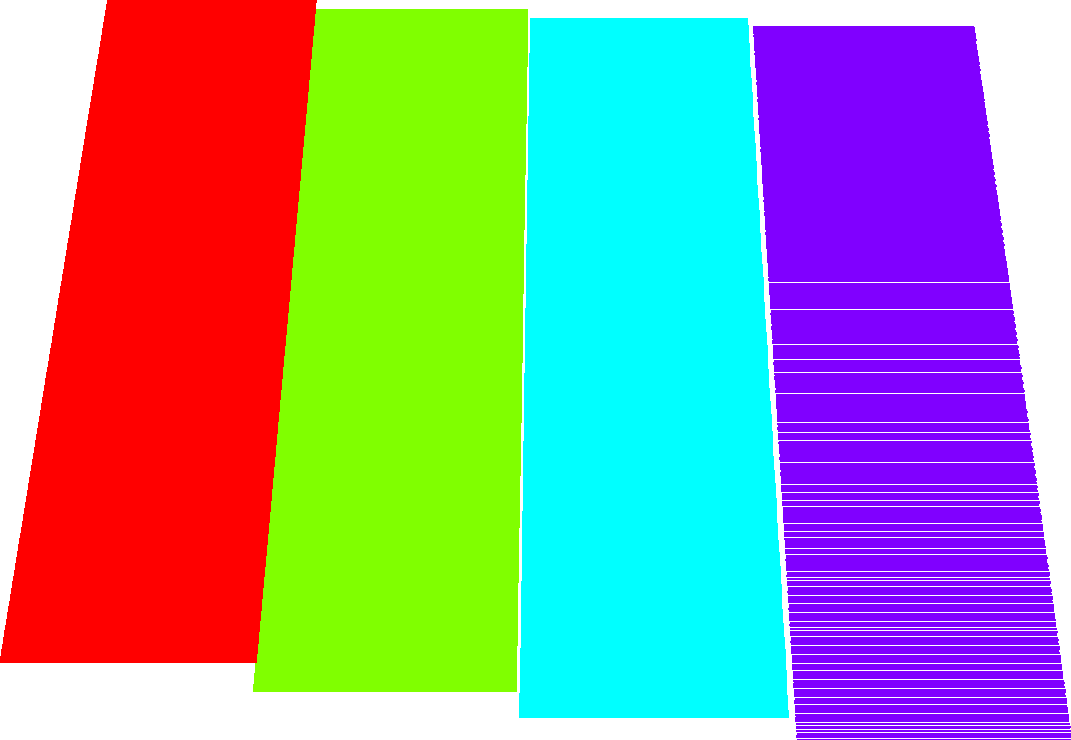}
        \caption{Our Planes}
    \end{subfigure}
    \caption{Depth image and plane mask of the 4 planes. From left to right, the frequency of the corrupting noise is $f=[0,2,10,100]$. Note that the colour of each plane follows the order of the clockwise HSV colour wheel, with red indicating the plane is maximally flat. }
    \label{fig:Synthatic_flatness}
\end{figure}

\begin{figure}
    \centering
    \begin{subfigure}{0.48\textwidth}
        \centering
        \includegraphics[width=\textwidth]{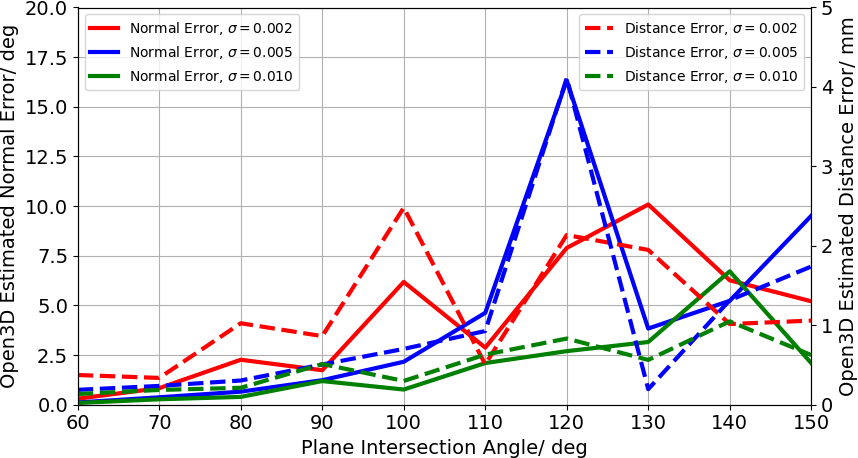}
    \end{subfigure}
    \begin{subfigure}{0.48\textwidth}
        \centering
        \includegraphics[width=\textwidth]{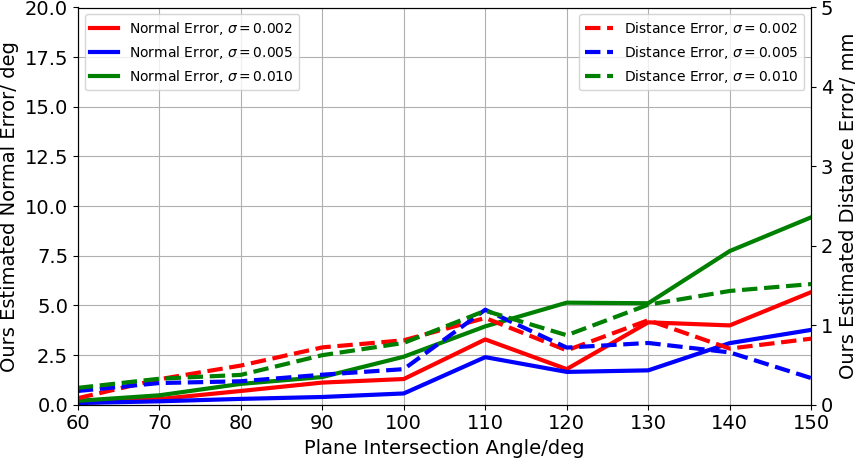}
    \end{subfigure}
    \caption{Graph of estimated normal and distance errors against plane intersection for both the Open3D and our approach. The scale of the x and y-axis are kept consistent between graphs.}
    \label{fig: Two_plane}
\end{figure}

Our proposed algorithm was tested against two synthetic depth images: one of a staircase similar to what was presented in \cite{yang2010plane}, and another with a tetrahedron protruding from a plane (See Fig \ref{fig:Synthatic_corner_staircase}). A zero-mean Gaussian noise with $\sigma=0.005$ was added to each image to simulate sensor noise. We applied both our proposed algorithm and the Open3D plane segmentation algorithm \cite{open3d} to the resulting point cloud at three different assumed noise levels, $\sigma=0.002,0.005,0.010$, searching for up to $N=8$ planes. Note that in the context of the Open3D algorithm, $\sigma$ refers to the chosen inlier threshold.

\begin{figure*}[t]
\captionsetup[subfigure]{labelformat = empty}
\newcommand{\rulesep}{\unskip\ \vrule\ }
    \centering
    \begin{subfigure}{0.16\textwidth}
        \centering
        \includegraphics[width=\textwidth]{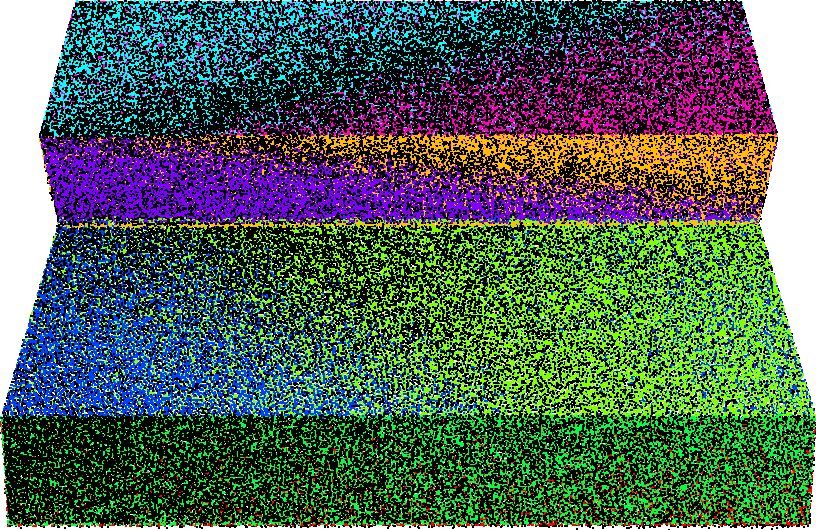}
    \end{subfigure}
    \begin{subfigure}{0.16\textwidth}
        \centering
        \includegraphics[width=\textwidth]{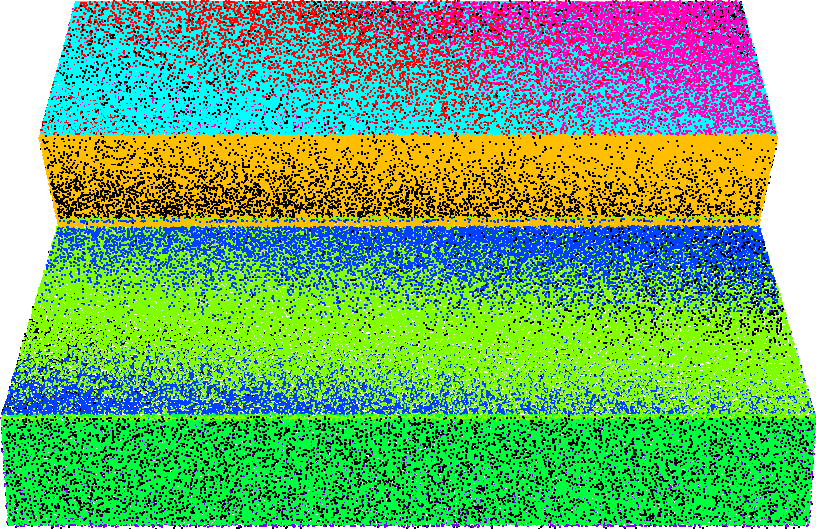}
    \end{subfigure}
    \begin{subfigure}{0.16\textwidth}
        \centering
        \includegraphics[width=\textwidth]{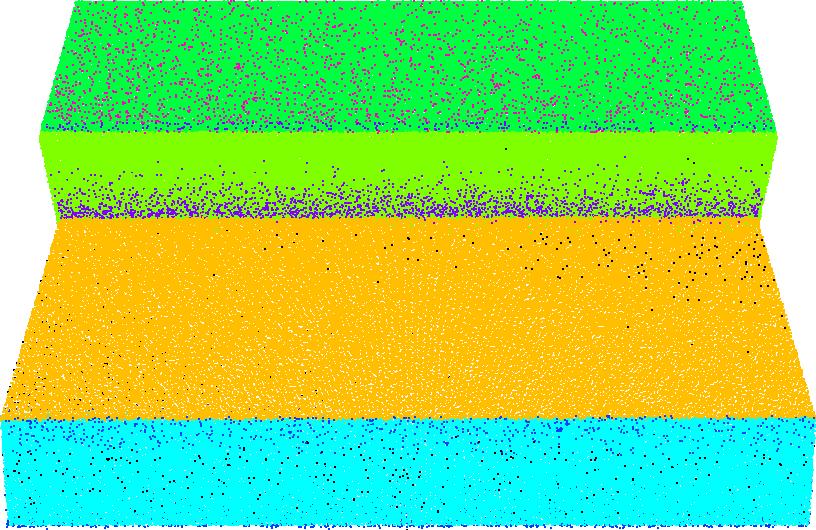}
    \end{subfigure}
    \rulesep
    \begin{subfigure}{0.16\textwidth}
        \centering
        \includegraphics[width=\textwidth]{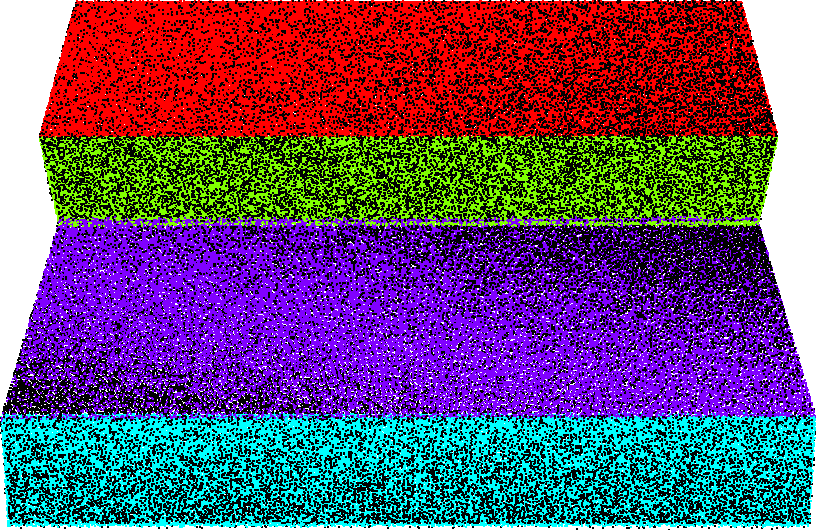}
    \end{subfigure}
    \begin{subfigure}{0.16\textwidth}
        \centering
        \includegraphics[width=\textwidth]{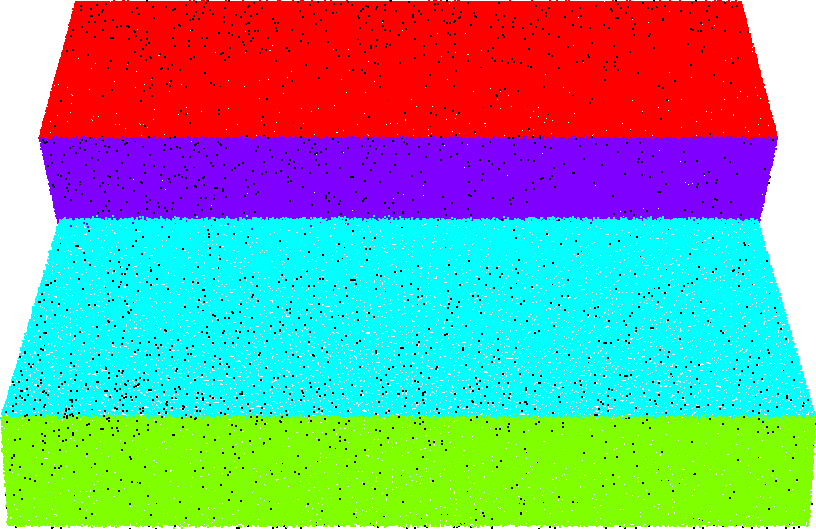}
    \end{subfigure}
    \begin{subfigure}{0.16\textwidth}
        \centering
        \includegraphics[width=\textwidth]{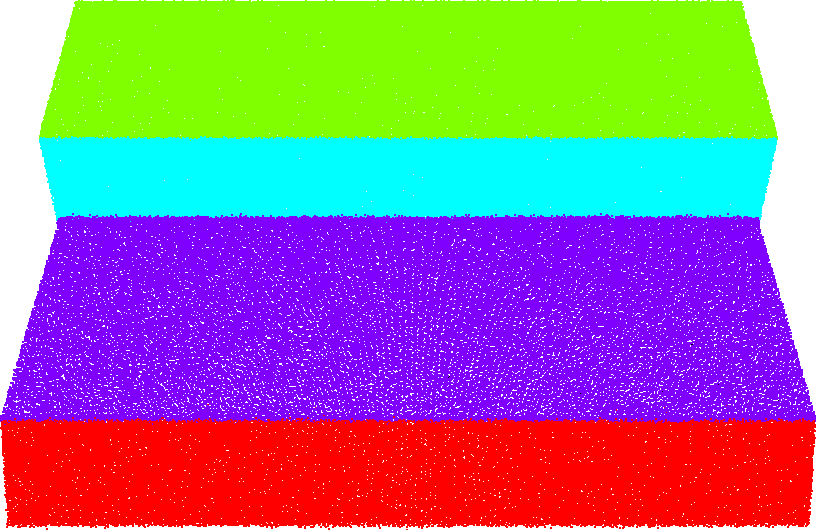}
    \end{subfigure}

    \begin{subfigure}{0.16\textwidth}
        \centering
        \includegraphics[width=\textwidth]{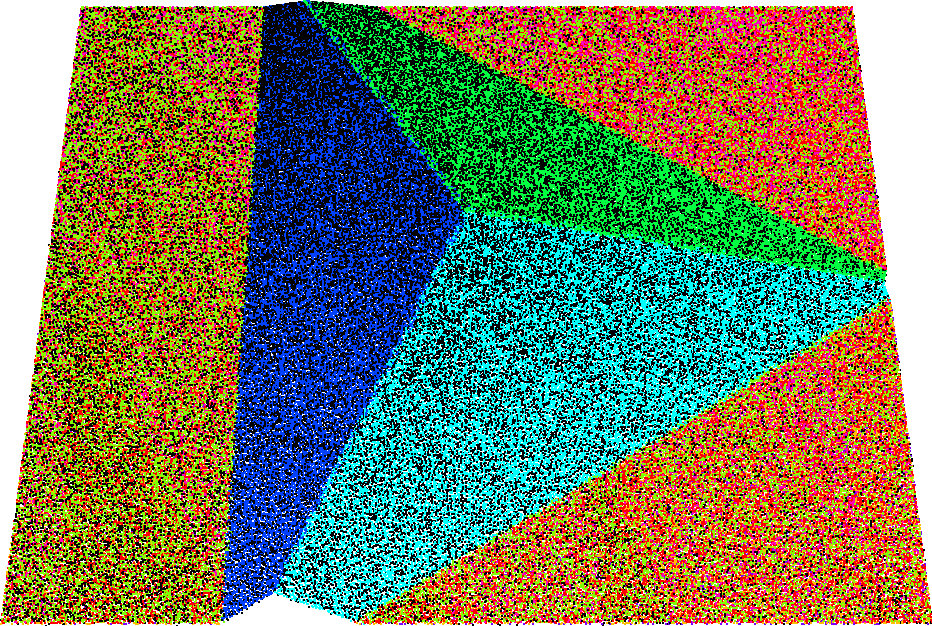}
        \caption{Open3D \\ $\sigma=0.002$}
    \end{subfigure}
    \begin{subfigure}{0.16\textwidth}
        \centering
        \includegraphics[width=\textwidth]{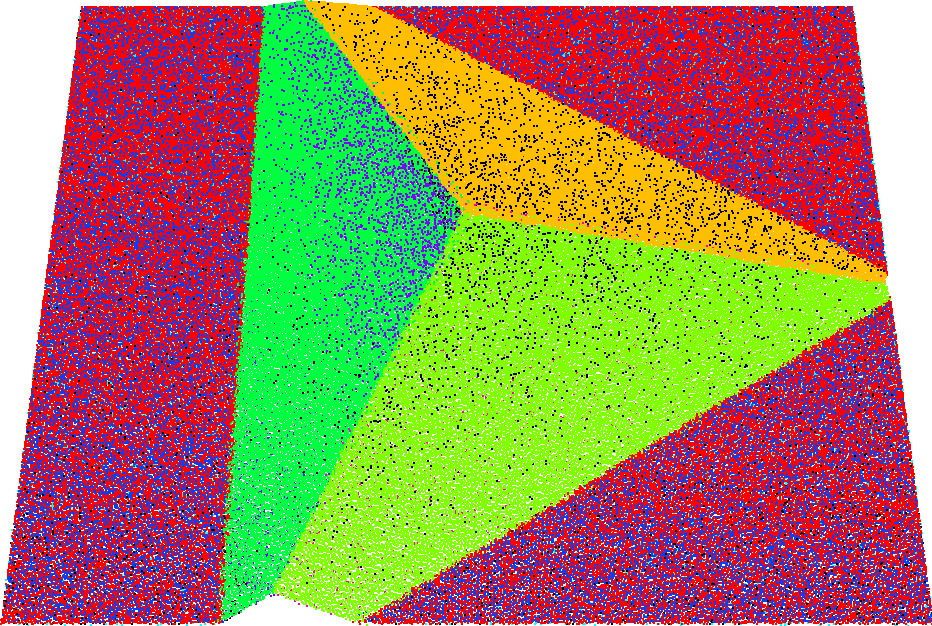}\
        \caption{Open3D \\ $\sigma=0.005$}
    \end{subfigure}
    \begin{subfigure}{0.16\textwidth}
        \centering
        \includegraphics[width=\textwidth]{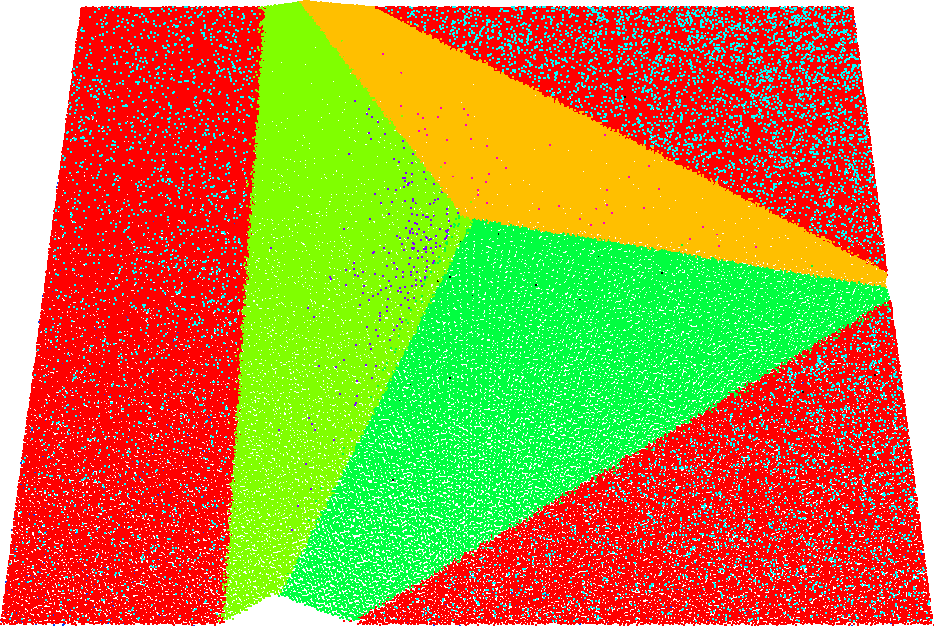}
        \caption{Open3D \\ $\sigma=0.010$}
    \end{subfigure}
    \rulesep
    \begin{subfigure}{0.16\textwidth}
        \centering
        \includegraphics[width=\textwidth]{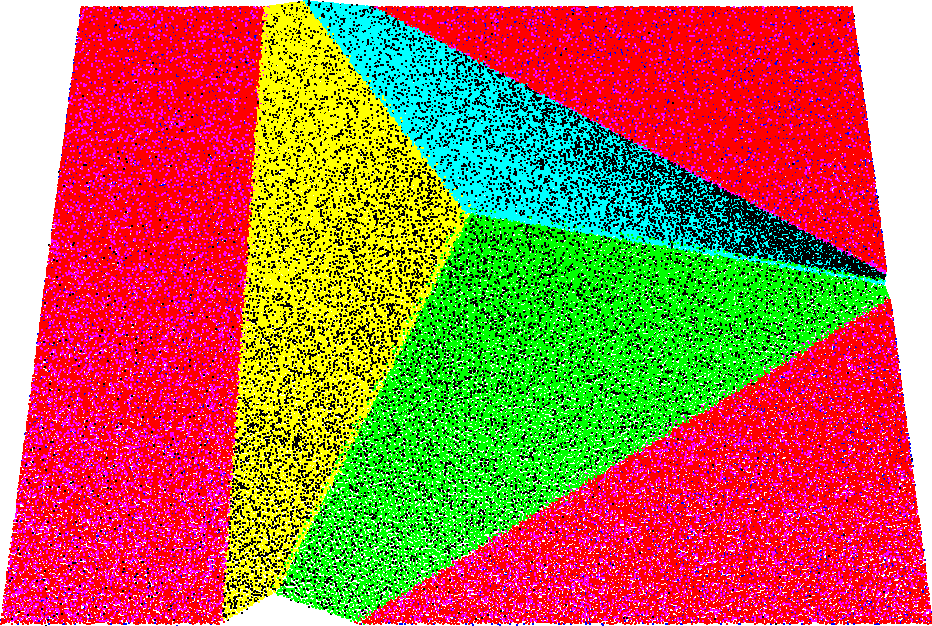}
        \caption{Ours \\ $\sigma=0.002$}
    \end{subfigure}
    \begin{subfigure}{0.16\textwidth}
        \centering
        \includegraphics[width=\textwidth]{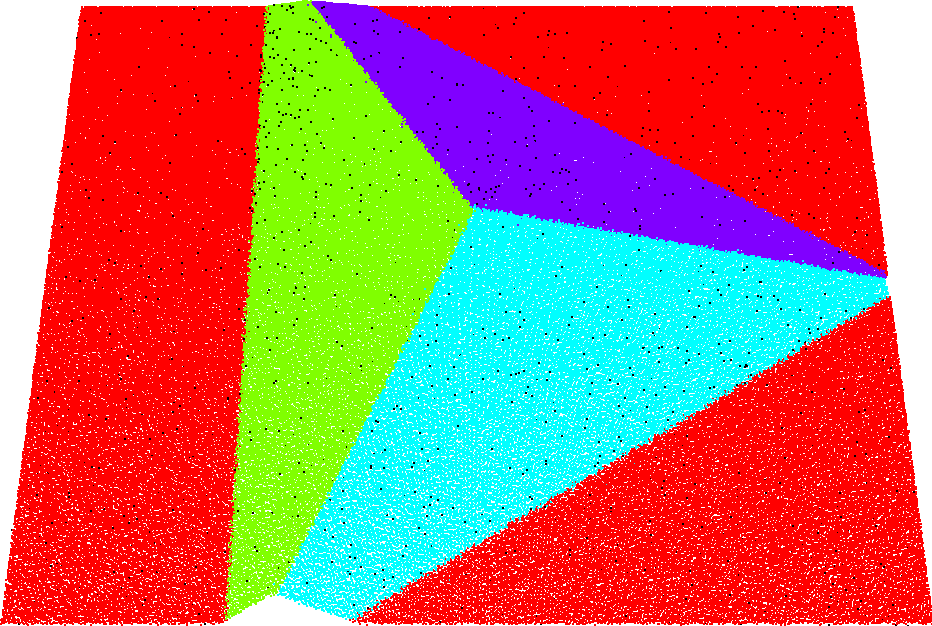}
        \caption{Ours \\ $\sigma=0.005$}
    \end{subfigure}
    \begin{subfigure}{0.16\textwidth}
        \centering
        \includegraphics[width=\textwidth]{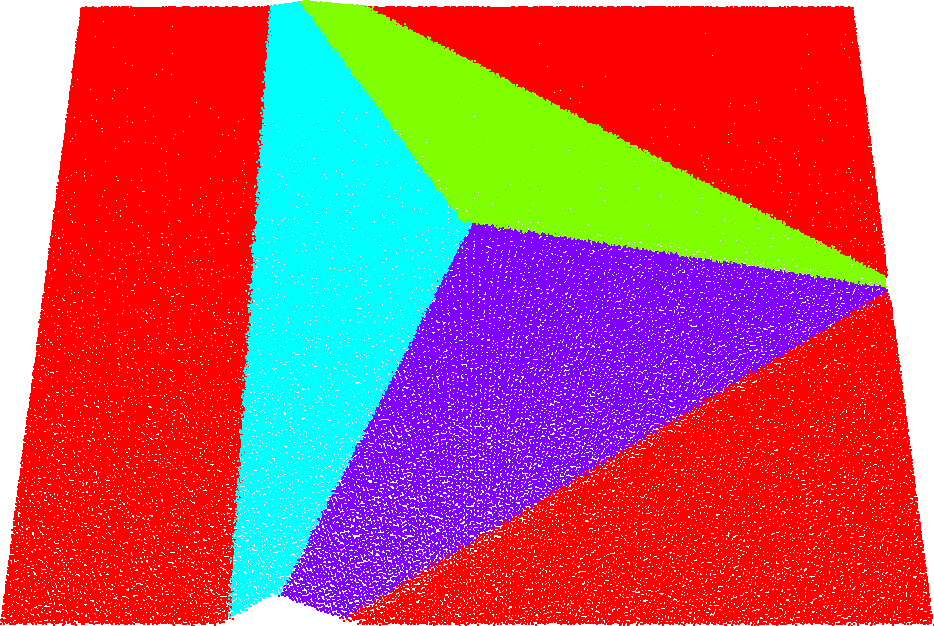}
        \caption{Ours \\ $\sigma=0.010$}
    \end{subfigure}

    \caption{Detected planes from Open3D vs our algorithm. Parameters used: $R=$depth.max$()-$depth.min(), $\epsilon=0.01$, $c=0.99$, $r=0.25$}
    \label{fig:Synthatic_results}
\end{figure*}

Figure \ref{fig:Synthatic_results} illustrates the detected planes from each experiment. We see that the Open3D approach always attempts to find eight planes of best fit, even after all four true planes are identified. This results in additional spurious planes being generated from the remaining noisy points, fragmenting the plane mask of our real planes. In contrast, our algorithm rejects any candidate plane found from the noisy points, as their information reduction is smaller than the increase in information due to model complexity. Hence, they are treated as outliers and only the real planes are kept. This effect is most noticeable when the assumed sensor noise matches the noise level of the depth image, emphasising the importance of selecting an appropriate sensor noise model in our approach. Note that a similar fragmentation effect can occur when the assumed noise level in our algorithm is lower than the actual sensor noise. When the sensor noise level was assumed to be $\sigma=0.002$, our algorithm detected an additional sparse plane (in purple) that is interlaced with the ground plane (in red) of the tetrahedron depth image. This is a consequence of our overly conservative assumption of the sensor noise, which overfits the previously excluded noisy points and thus misidentifies the ground truth model. 

Beyond preventing false positive plane detection, our algorithm also ranks the relative fit of each detected plane. We demonstrated this by applying both algorithms to a synthetic depth image containing four equally sized planes (See Fig \ref{fig:Synthatic_flatness}). Three of the planes were corrupted by a sinusoidal wave along the width of the image, each at a different frequency $f$. Intuitively, as $f$ increases, the quality of the plane should degrade. Setting $\sigma$ as the wave amplitude, we apply both approaches to the depth image. It is observed that the Open3D approach fails to detect the individual planes correctly, with the detected planes interlaced between the peaks and troughs of the sinusoidal noise. In contrast, our algorithm successfully detected each individual plane and ordered them according to the frequency of the corrupting noise, proving another key strength in our approach.

To better quantify the accuracy of our estimated plane parameters, both algorithms were applied to a series of depth images comprising two intersecting planes with varying angles of intersection. As before, a Gaussian sensor noise with $\sigma=0.005$ was added to each image and both algorithms were tested at the three different assumed noise levels. Figure \ref{fig: Two_plane} graphs the normal and distance error of the extracted plane parameters against the angle of intersection between the planes. Similar to our earlier experiments, the estimated plane parameters from our method are the most accurate when $\sigma$ matches the sensor noise. Additionally, both errors worsen as the angle of intersection increases. This is because a larger intersection angle reduces the relative normal angle between the normals of the two planes, making it harder for our algorithm to differentiate points from each plane. Compared to the plane parameters generated from the Open3D approach, the errors from our algorithm are smaller and increase slower with the intersection angle, demonstrating the strengths of our approach.

\subsection{Real-world Data: NYU-v2}
\captionsetup[subfigure]{labelformat = empty}
\newcommand{\heighta}{2.1cm}
\begin{figure*}[htbp]
    \centering
    \begin{subfigure}{0.16\textwidth}
        \centering
        \includegraphics[height=\heighta]{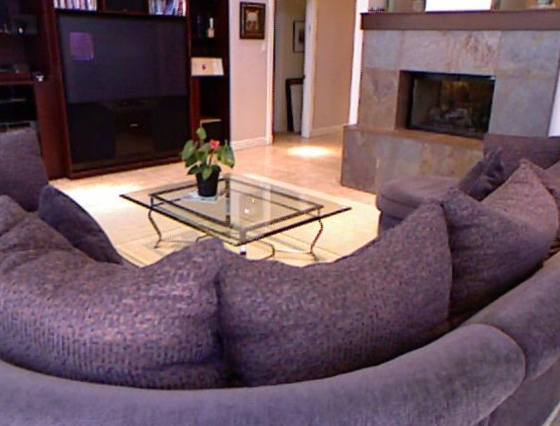}
    \end{subfigure}
    \begin{subfigure}{0.16\textwidth}
        \centering
        \includegraphics[height=\heighta]{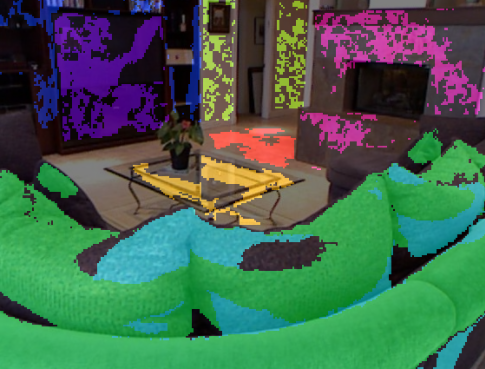}
    \end{subfigure}
    \begin{subfigure}{0.16\textwidth}
        \centering
        \includegraphics[height=\heighta]{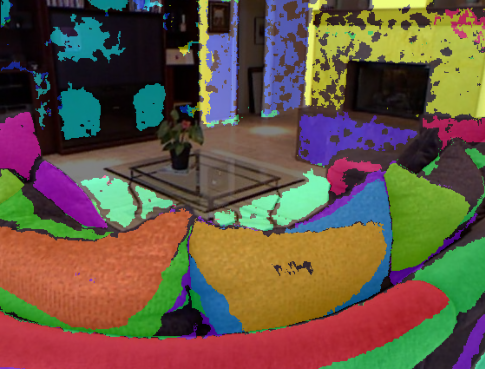}
    \end{subfigure}
    \begin{subfigure}{0.16\textwidth}
        \centering
        \includegraphics[height=\heighta]{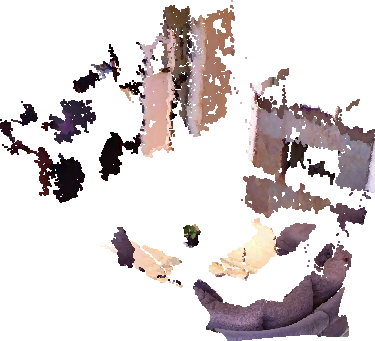}
    \end{subfigure}
    \begin{subfigure}{0.16\textwidth}
        \centering
        \includegraphics[height=\heighta]{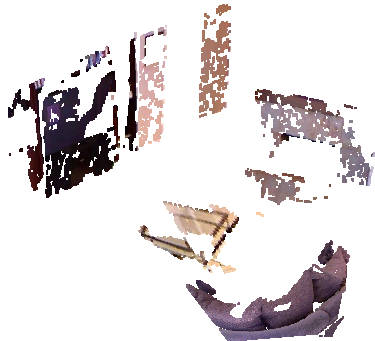}
    \end{subfigure}
    \begin{subfigure}{0.16\textwidth}
        \centering
        \includegraphics[height=\heighta]{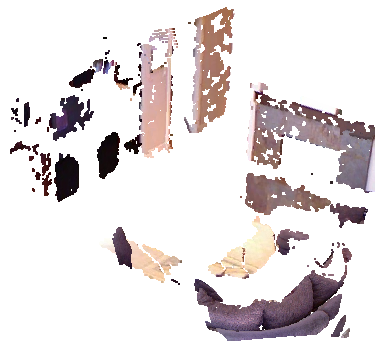}
    \end{subfigure}
    \hspace{5mm}
    \begin{subfigure}{0.16\textwidth}
        \centering
        \includegraphics[height=\heighta]{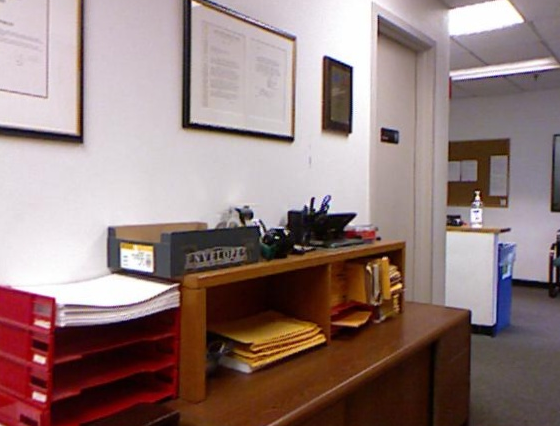}
    \end{subfigure}
    \begin{subfigure}{0.16\textwidth}
        \centering
        \includegraphics[height=\heighta]{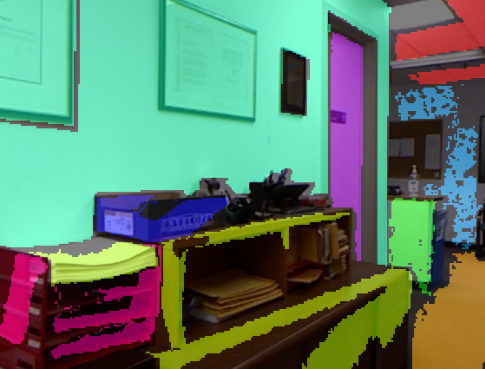}
    \end{subfigure}
    \begin{subfigure}{0.16\textwidth}
        \centering
        \includegraphics[height=\heighta]{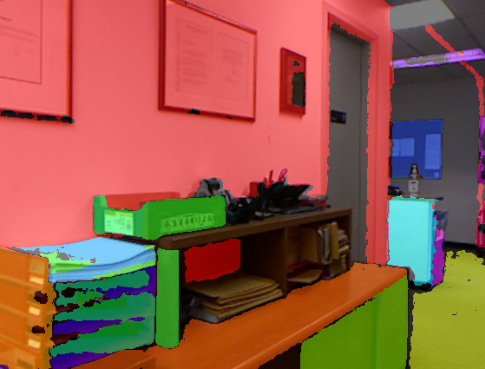}
    \end{subfigure}
    \begin{subfigure}{0.16\textwidth}
        \centering
        \includegraphics[height=\heighta]{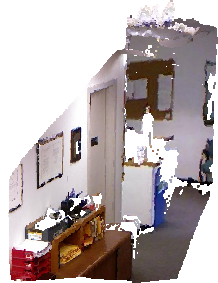}
    \end{subfigure}
    \begin{subfigure}{0.16\textwidth}
        \centering
        \includegraphics[height=\heighta]{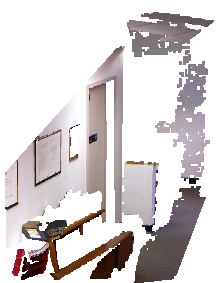}
    \end{subfigure}
    \begin{subfigure}{0.16\textwidth}
        \centering
        \includegraphics[height=\heighta]{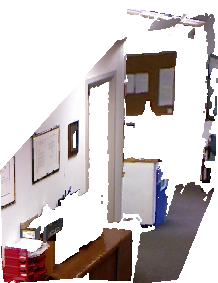}
    \end{subfigure}
    \begin{subfigure}{0.16\textwidth}
        \centering
        \includegraphics[height=\heighta]{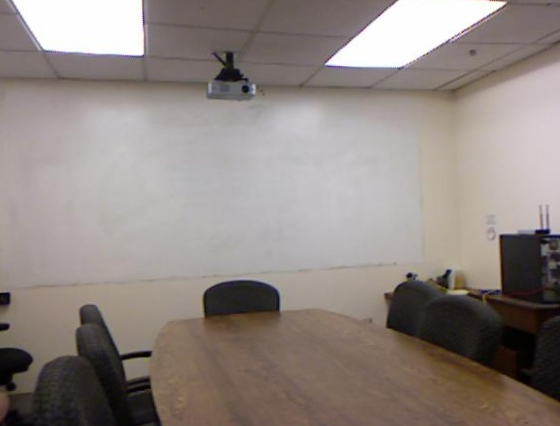}
        \caption{RGB Image}
    \end{subfigure}
    \begin{subfigure}{0.16\textwidth}
        \centering
        \includegraphics[height=\heighta]{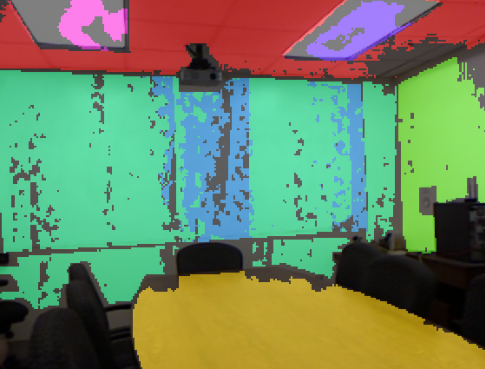}
        \caption{GT Plane Mask}
    \end{subfigure}
    \begin{subfigure}{0.16\textwidth}
        \centering
        \includegraphics[height=\heighta]{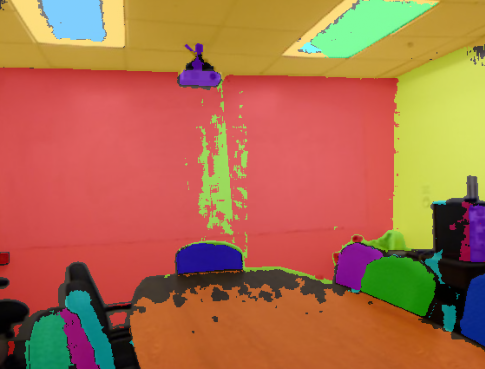}
        \caption{Our Mask}
    \end{subfigure}
    \begin{subfigure}{0.16\textwidth}
        \centering
        \includegraphics[width=\textwidth]{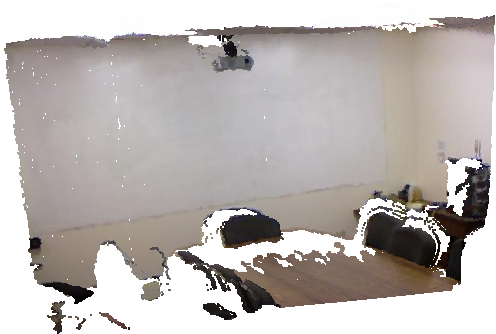}
        \caption{Depth Points}
    \end{subfigure}
    \begin{subfigure}{0.16\textwidth}
        \centering
        \includegraphics[width=\textwidth]{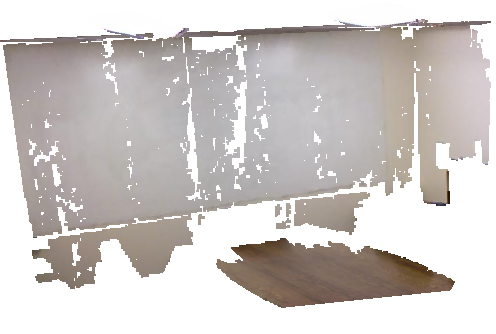}
        \caption{GT Planes}
    \end{subfigure}
    \begin{subfigure}{0.16\textwidth}
        \centering
        \includegraphics[width=\textwidth]{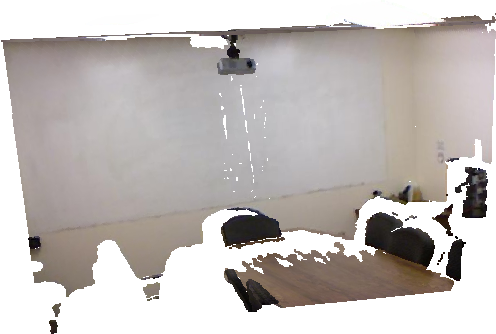}
        \caption{Our Planes}
    \end{subfigure}

    \caption{Comparison of the ground truth plane mask against that from our algorithm. Since the ground truth plane parameters only included the normal vector of each plane, the plane distance was estimated to be the average of distance of each assigned points. Due to limitations on the HSV colour space, only the 16 best planes are depicted in each our segmented plane mask.}
    \label{fig:compare_visual}
\end{figure*}
To evaluate how well our algorithm performs under real-world conditions, we decided to test our algorithm on the NYU-v2 dataset \cite{nyu_v2_dataset}. This consists of 1449 pairs of RGB and depth images, captured from a diverse set of indoor environments using an XBox Kinect. We have decided to compare our detected planes with the segmented planes from \cite{associative_embedding}. These planes have been accepted as the ground truth by many other plane estimation neural network models, which rely on these planes as both training and evaluation data \cite{PlaneTR}\cite{PlaneRecTR++}\cite{Zhang2023PlaneSeg}. This dataset was hosted by \cite{shi2023planerectr}, which also provided an evaluation script that allowed us to benchmark our segmented planes based on three popular segmentation metrics: Variation of Information (VOI), Rand Index (RI), and Segmentation Covering (SC). 

For our experiments, we first employ the default Segment-Anything model \cite{kirillov2023segany} to generate a list of partitions with a stability score of at least $0.98$. Our proposed algorithm was then applied to each partitioned region using the following parameters: maximum number of planes $N=4$, confidence level $c=0.99$ and inlier ratio $r=0.2$. Taking the inverse of the summation of all partition masks, we were left with a final partition of all excluded points. The algorithm was also applied to this region using the same parameters, with the maximum number of planes increased to $N=8$ and inlier ratio reduced to $r=0.1$. Finally, the detected planes were ranked according to their information reduction, with the smaller planes merged if they provide a negative total information reduction for a larger plane. Note that since many smaller planes exist within each semantic partition, we produced a larger number of planes compared to the ground truth. Hence, only the $k$ best planes, with the most negative model information reduction, were selected for evaluation in each scene, where $k$ matches the number of planes found in the ground truth. We repeated our experiments using two different $\sigma$ functions. First, an experimentally derived model that estimated the sensor noise of the Kinect as $\sigma(z)=0.0012+0.0019*(z-0.4)^2 $ m \cite{kinect_noise}. Next, a proportional noise model with $\sigma=0.01*z$ m that was more representative of what is typically reported in the specification sheets of depth cameras \cite{gemini2xl}\cite{d435}. 

\begin{table}
\centering
\begin{tabular}{p{15em}|ccc}
\hline
\multirow{2}{*}{Methods} & \multicolumn{3}{c}{NYUv2-Plane} \\
& VOI ($\downarrow$) & RI ($\uparrow$) & SC ($\uparrow$) \\
\hline
PlaneTR \cite{PlaneTR} & 1.110 & 0.898 & 0.726 \\
PlaneRecTR \cite{shi2023planerectr} \cite{PlaneRecTR++} & 1.045 & 0.915 & 0.745 \\
PlaneSeg \cite{Zhang2023PlaneSeg} & 1.684 & 0.856 & 0.741 \\
\hline
Ours (Experimental) & 1.259 & 0.890  & 0.703 \\
Ours (Proportional) & \textbf{0.874} & \textbf{0.934} & \textbf{0.799} \\
\hline
\end{tabular}
\caption{ Comparison of plane segmentation results from our algorithm against other neural network-based methods}
\label{Table 1: Metrics}
\end{table}

Table \ref{Table 1: Metrics} shows the metrics of our segmented planes compared to neural network-based methods. As expected, our approach performs significantly better since it is applied directly to the depth images, whereas the neural networks must infer the planes through an RGB input. Interestingly, our simplified proportional noise model outperformed the more complex experimental noise model in every metric. This is noteworthy as the experimental noise model is more conservative and may be underestimating the true sensor noise. We therefore present some of the planes generated from our algorithm using the proportional noise model in Fig \ref{fig:compare_visual} to perform a qualitative evaluation. The first scene depicted had received the worst metric across all three categories. This is partly because the couch, a dominant feature in the scene, was wrongly segmented into two planes. In comparison, our algorithm is able to break it down into its component planes, providing a much more accurate segmentation. A similar result can be seen in the second scene, where a false plane was predicted to run diagonally along the table. Our method also reduces the problem of plane fragmentation, as evident in the third image, where the wall was grouped into a more complete plane. These anecdotal examples help illustrate the many limitations that exist within the current plane dataset and demonstrate how our algorithm can help generate more reliable ground truth planes for model training.

Due to the large search space of possible plane candidates, our algorithm suffers a relatively slow runtime. Including the segmentation step using SAM, our Python implementation averages 308 seconds to run for each frame using a Intel Core I3-10100F and Nvidia GeForce RTX 3070. We also provide a slightly faster C++ implementation that reduces the runtime to 21 seconds per frame; however, this version does not include integration with RGB segmentation.

\subsection{Limitations: Intersection Ambiguity}
\label{Section: Limitations}
\begin{figure}[b!]
    \centering
    \begin{subfigure}{0.15\textwidth}
        \centering
        \includegraphics[width=\textwidth]{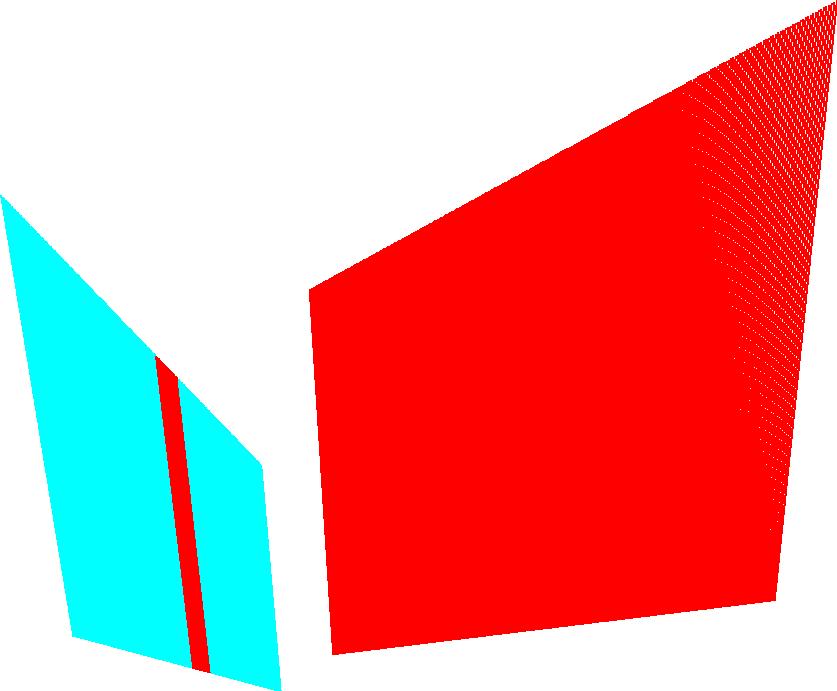}
        \caption{No Post-Processing}
    \end{subfigure}
    \begin{subfigure}{0.15\textwidth}
        \centering
        \includegraphics[width=\textwidth]{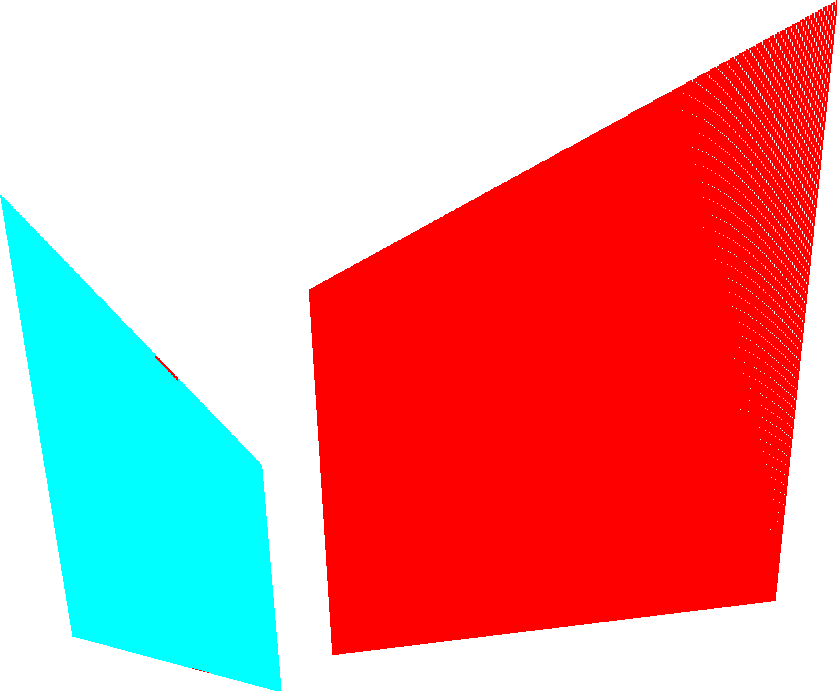}
        \caption{Post-Processing}
    \end{subfigure}
    \begin{subfigure}{0.15\textwidth}
        \centering
        \includegraphics[width=\textwidth]{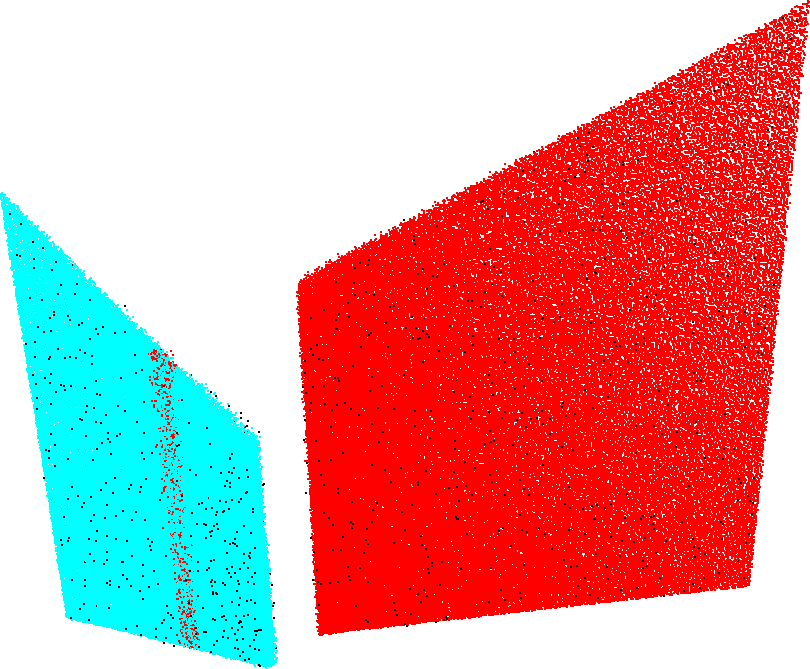}
        \caption{With Noise}
    \end{subfigure}
    \caption{Points along the intersection of two planes may be misassigned by our algorithm. By comparing the surface normal of each plane, the plane mask can be corrected by the post-processing step. However, the efficacy of this solution diminishes with the inclusion of sensor noise, which corrupts our normal estimates, resulting in noisy patches of misassigned masks.}
    \label{fig:limitation}
\end{figure}
Despite the strengths of our algorithm, it struggles with mask assignment ambiguity near plane intersections, similar to other random sub-sampling methods. For example, consider the depth map of a T-junction viewed from a top-down perspective, a large flat plane is observed together with a second, disconnected plane further away. Points along the line of intersection belong to both planes, but for most practical applications, it is preferable that they be assigned solely to the front one. This, however, is not achievable within the current framework of our algorithm, as it is focused solely on information minimisation and ignores the overall structure of the point cloud.

To mitigate this problem, we experimented with a post-processing step that reassigns each pixel based on their surface normal and distance. By comparing the angle and distance error of each pixel against all found plane parameters, the pixel is assigned to the most similar plane that still provides a negative contribution to the overall information. To estimate the surface normal of each pixel, its surface tangent vectors along both the x- and y-axes are first calculated using a local gradient kernel, whose cross product gives the surface normal vector. This is then smoothed with a Gaussian filter. The effects of the pixel reassignment can be seen in Fig \ref{fig:limitation}, where the previously misassigned strip of points along the line of intersection are now correctly reassigned to the top plane.

Unfortunately, the effectiveness of this post-processing step diminishes in the presence of sensor noise, which corrupts our surface normal estimates. As a result, the reassignment process becomes inaccurate, leaving behind noisy patches that retain the wrong plane mask. Increasing both the gradient kernel and Gaussian filter can mitigate this problem, but it also destroys local information. This can lead to a reduction in plane parameter accuracy and the merger of close but yet distinct planes in real-world data. For these reasons, this solution was not included within our algorithm and left instead as an optional step, useful for certain applications.

\section{Conclusion}

In summary, our algorithm effectively replaces the arbitrary inlier threshold of the conventional RANSAC method with a theoretically rigorous assignment criterion based on model information. This reformulates the plane detection problem as one of information optimisation, helping to prevent false positive plane detections by identifying the most likely model. By incorporating the physics and noise characteristics of the depth sensor, our approach provides a simple framework for accurate plane detection that requires minimal manual tuning. Additionally, the information reduction of each plane serves as a convenient metric that ranks the quality of our detected planes. We have demonstrated the effectiveness of our algorithm using synthetic data, showing measurable improvements in plane parameter accuracy compared to the default RANSAC plane segmentation of Open3D. Our algorithm also successfully rectifies errors typically encountered in neural network-based plane detection methods in real-world data. This approach may be applied to robotics, enabling more robust and accurate plane detection in dynamic environments.

Moving forward, we aim to enhance the robustness of the algorithm by estimating all plane parameters concurrently, potentially using methods like MultiRANSAC \cite{multiransac}. While this approach can become computationally intractable when the inlier ratio for each plane is low and the assumed maximum number of planes is large, we believe some of these challenges can be addressed by integrating localised plane estimation methods into our existing framework. We hope to determine if such an approach is effective in our future work.

\balance
\bibliographystyle{IEEEtran}
\bibliography{intro,references,background}

\end{document}